\documentclass[runningheads]{llncs}
\usepackage[pdftex]{graphicx}

\usepackage{amsmath}
\usepackage{bm}
\usepackage{cite}
\usepackage{subfig}
\usepackage{hyperref}

\begin{document}
\title{Interpretation of ResNet by Visualization of Preferred Stimulus in Receptive Fields}

\author{
Genta Kobayashi\inst{1} \and
Hayaru Shouno\inst{1}
}

\institute{
The University of Electro-Communications, Chofu, Tokyo, Japan\\
\email{\{genta-kobayashi,shouno\}@uec.ac.jp}
}
\maketitle              % typeset the header of the contribution
\begin{abstract}
One of the methods used in image recognition is the Deep Convolutional Neural Network (DCNN).
DCNN is a model in which the expressive power of features is greatly improved by deepening the hidden layer of CNN. 
The architecture of CNNs is determined based on a model of the visual cortex of mammals.
There is a model called Residual Network (ResNet) that has a skip connection.
ResNet is an advanced model in terms of the learning method, but it has not been interpreted from a biological viewpoint.
In this research, we investigate the receptive fields of a ResNet on the classification task in ImageNet.
We find that ResNet has orientation selective neurons and double opponent color neurons.
In addition, we suggest that some inactive neurons in the first layer of ResNet affect the classification task.

\end{abstract}

\section{Introduction}
% history
In this decade, deep convolutional neural networks (DCNNs) have been used in many areas such as image processing, audio signal processing, language processing, and so on.
Especially, in image classification task, DCNN showed higher performance rather than that of the previous works in the field of computer vision\cite{ILSVRC15}.
DCNN is a model in which the expressive power of features is greatly improved by deepening the hidden layer of the convolutional neural network (CNN). 
Characteristics of CNN are build to hierarchically stack convolutional layers and pooling layers.
Both architectures are determined based on simple cells and complex cells that are the visual cortex of mammals\cite{Fukushima1980}.
CNN are added constraints from a biological point of view e.g. weight sharing and sparse activation.
LeCun \textit{et al}. \cite{Lecun1998} propose a model of CNN called LeNet-5 for the classification task of digit images, and apply the backpropagation algorithm of the gradient learning method to the model.
Krizhevsky \textit{et al}. \cite{AlexNet2012} show the effectiveness DCNN on the natural image classification task.
In the wake of their achievements, many researchers proposed various deep models \cite{simonyan2014vgg, szegedy2015inception}.
He \textit{et al}. \cite{He2016ResNet} also proposed a DCNN model called residual network (ResNet) that has skip connections for bypassing the layers. The ResNet improves the performance of the visual classification task drastically.

The success of DCNNs accelerated the need of understanding them from multiple angles.
From the viewpoint of neuroscience, 
Yamins \textit{et al}. experimentally showed the similarity between the visual cortex of the primate and a DCNN trained for classification task\cite{Yamins8619}.
On the other hand, from an engineering viewpoint, the mainstream method of understanding DCNN is based on visualization of the inner expression of DCNNs using the gradient backward projection \cite{simonyan2013deep, selvaraju2017grad, springenberg2014striving}.
These methods use the differentiability of the function of DCNNs in the task.

The basic structure of the DCNNs is based on the inspiration from the biological viewpoint\cite{Fukushima1980}, however, non-biological improvements, which have been proposed in these years, increases the interpretation difficulties. 
For instance, ResNet is an improved model so that the gradient based learning methods work well. 
To understand ResNet, Liao \& Poggio study the relation between a model of ResNet and the visual cortex\cite{liao2016bridging}.
They use that the model of ResNet is similar to recurrent neural networks that had a feedback connection.
The study shows the relationship between a model of ResNet and recurrent neural network,
and then between the ventral stream and the model stacked recurrent neural network.
However the model is added a strong constraint and is not commonly used.

In this research, in order to understand ResNet, we focus it from the viewpoint of the development of the preferred stimulus in receptive fields under the visual scene classification task with ImageNet\cite{imagenet_cvpr09, ILSVRC15}.
The receptive field is a basic concept of the visual cortex system. Roughly speaking, it means the part of the visual input area in which a neuron is able to respond. The preferred stimuli make the strong response of the neuron. 
We try to use the idea of the preferred stimulus in the receptive field to reveal properties of the ResNet.

\section{Methods}
\subsection{Residual Network}
\begin{figure}[tbp]
    \centering
    \includegraphics[width=0.9\textwidth]{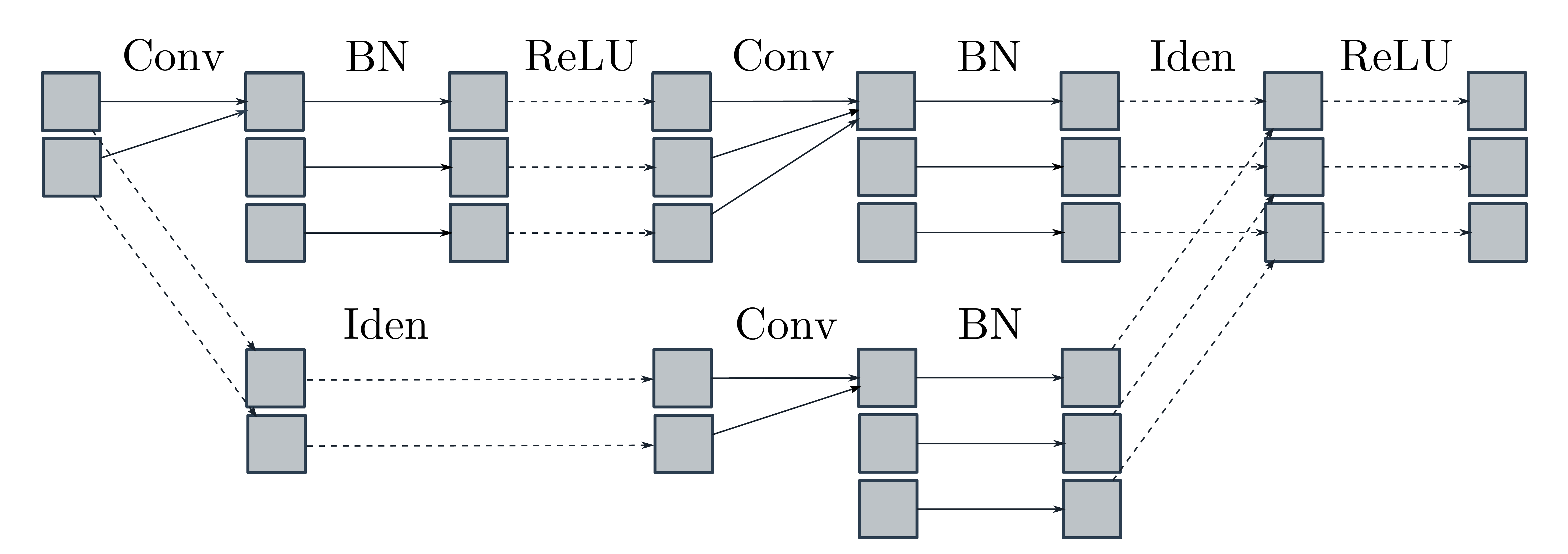}
    \caption{
    Schematic diagram of the ResNet34: Each rectangle represents the feature map. The fixed arrows show the connectivity with trainable, and dashed ones show the connectivity with a fixed weight.
    Conv: convolutinonal layer, BN: batch normalization, ReLU: ReLU function as $\max(x, 0)$, Iden: identity function.
    }
    \label{fig:downsampling_block}
\end{figure}

He \textit{et al}. proposed the concept of Residual Network (ResNet) and showed several models of ResNet, \textit{e.g.} ResNet18, ResNet34, ResNet50, ResNet101, and ResNet152\cite{He2016ResNet}.
ResNet contains characteristic architecture called ``skip connection'' or ``residual connection''.
The concept of the residual connection is to divide the mapping function into linear and non-linear parts explicitly.
Let an input vector as $\bm{x}$, the output vector as $\bm{y}$, and nonlinear part of mapping function as $F(\cdot)$. Then skip connection is represented as:
\begin{equation}
    \bm{y} = \bm{x} + F(\bm{x}).
    \label{eq:residual}
\end{equation}
When the dimensions of $\bm{x}$ and $F(\bm{x})$ are different, $\bm{x}$ is mapped to sum them by a mapping function.
The original ResNets introduce a down-sampling block contains a convolutional layer at some skip connections.
Fig. \ref{fig:downsampling_block} shows the schematic diagram of the components of the ResNet called Residual block.
In order to treat skip connection in the Residual block, we introduce pseudo feature maps for the identical part of eq.\eqref{eq:residual}.
In the figure, each rectangle shows the feature map, the fixed arrows show the connectivity with trainable weights, and the dashed ones show the connectivity with a fixed weight. 
We also introduce named PlainNet as the model excluding all the skip connections for comparison. 
We use ResNet34 and PlainNet34 for our experiment since the ResNet34 shows higher performance rather than those of the other ResNets models and previous DCNNs in our preliminary experiments.

\subsection{Receptive Field}
In the context of biological visual systems, the receptive field is the area on the retina to which a neuron has responded.
It is considered that the receptive field contains the center and the surround area.
Hubel \& Wiesel shows almost all the receptive fields in the early visual cortex are very small\cite{hubel1959receptive}, and they become large as the hierarchy deepens.
Their work inspires the Neocognitron\cite{Fukushima1980}, which is one of the origin of the DCNN, and influences many image recognition researches.

\begin{figure}[tbp]
    \centering
    \includegraphics[width=0.45\textwidth]{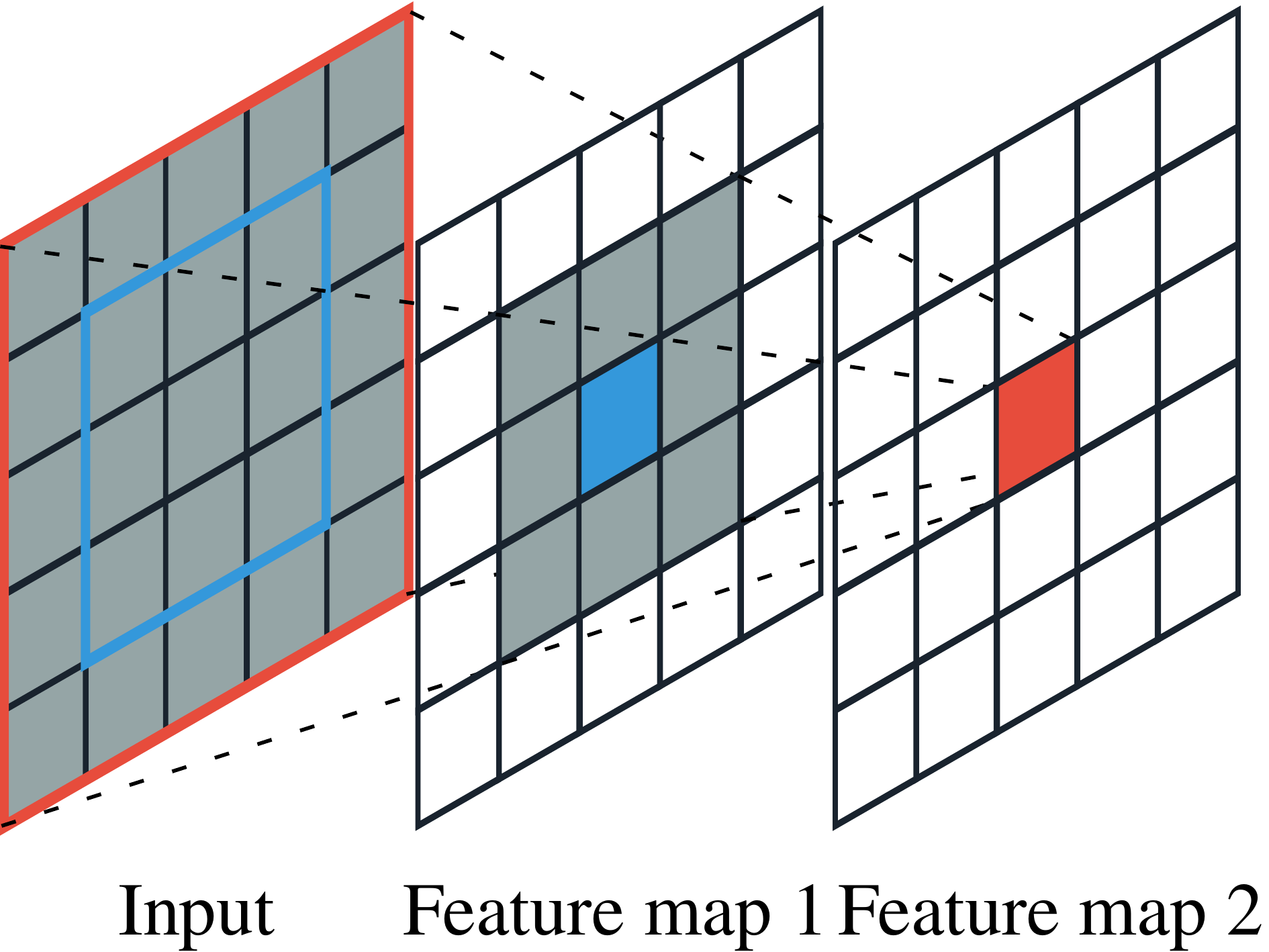}
    \caption{Overview of receptive field.
             Each black boder rectangle is a neuron.
             The area inside the blue border on input is the receptive field and corresponds to the blue neuron in feature map $1$.
             The area inside the red border on input is the receptive field and corresponds to the red neuron in feature map $2$.
             }
    \label{fig:rf_show}
\end{figure}

In the context of CNN, each neuron has the receptive field and also has preferred stimuli that are a part of the patch in the input image.
Fig. \ref{fig:rf_show} shows an overview of the receptive field.
The most right rectangle shows the feature map of the focused layer, and the middle and the left one shows the intermediate feature map and input respectively. The feature map has neurons aligned with 2-dimensional lattice. When we choose a neuron in the focused feature map, we can determine the connected area in the middle and the input. 
Thus, the preferred stimuli for the focused neuron are appeared in the red rectangle.
Zeiler \textit{et al}. \cite{zeiler2014visualizing} show samples of the preferred stimuli of DCNN and report the characteristic of each layer.
Showing its sample is a simple method to understand trained features of CNN.
We use the preferred stimulus to investigate the characteristic of neurons in this research.
Let $\bm{x}$ be an image of $H \times W$ then the receptive field is a set of the spatial index.
We can formally describe the receptive field image on the receptive field $\bm{r}$ of the image $\bm{x}$ as $\bm{x}[\bm{r}]$.

\subsection{Visualization by Using Gradient}

Many researchers use gradient base visualization methods to understand deep neural networks\cite{erhan2009visualizing, simonyan2013deep, selvaraju2017grad, springenberg2014striving}.
First the work is activation maximization of Erhan \textit{et al}. \cite{erhan2009visualizing} and Simonyan \textit{et al}. \cite{simonyan2013deep} apply it to DCNN.
Activation maximization is to calculate an input that maximizes the activation of the neuron as an optimization problem.
Let $\bm{\theta}$ denote parameters of neural network and let $f(\bm{\theta}, \bm{x})$ be the activation of a neuron on a given input $\bm{x}$.
Assuming a fixed $\bm{\theta}$, the method is represented as
\begin{equation}
    \bm{x}^{*} = \arg \max_{\bm{x}} \{ f(\bm{\theta}, \bm{x}) - \lambda \| \bm{x}{\|}_2 \}.
    \label{equ:act_max}
\end{equation}
Since the solution we are interested in is a direction of input space, we add $L_2$ norm constraint and a regularisation parameter $\lambda$. 
In general, this method is solved by gradient ascent of iterative methods because this is a non-convex optimization problem.
This method can be applied to any differentiable models but the resulting solution may be a boring local solution.

\section{Experiment and Results}
\subsection{Training ResNets}
We train ResNet34 and PlainNet34 with ImageNet dataset in the manner of He \textit{et al}. \cite{He2016ResNet} and Szegedy \textit{et al}. \cite{szegedy2015inception}.
The images in ImageNet have $3$ color channels and are whitening with the channels.
We apply the stochastic gradient descent method with an initial learning rate of $0.01$, a momentum of $0.9$, and use a weight decay of  $10^{-4}$.
The learning rate is divided by $10$ every $30$ epochs.
The total training epoch is $90$ with mini-batch size $256$.
In the training, the input images of $224 \times 224$ size are randomly resized by an area scale between $8\%$ and $100\%$, and whose aspect ratio is chosen randomly between $3 / 4$ and $4 / 3$.

\subsection{Visualization Filters}
\begin{figure}[tbp]
    \centering
    \subfloat[Virtual filter $\widetilde{W}^{2}_{2}$ and
            sorted the weight values.
            The right graph show the values of weight $W^{2}_{2k \Tilde{i}_{2k} \Tilde{j}_{2k}}$
            and sorted index of x-axis is an index $k$ sorted in descending order.]{
            \includegraphics[width=\textwidth]{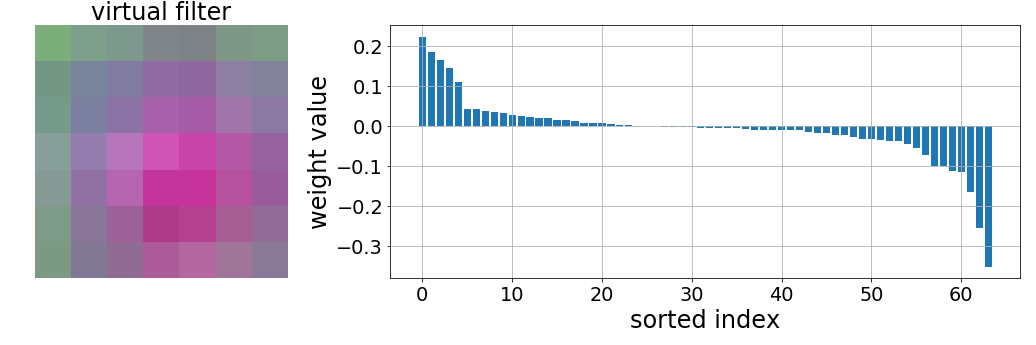}\label{fig:vfilter_downsampling_block_a}}
    \\
    \centering
    \subfloat[Filters $W^{1}$ sorted by weight $W^{2}_{2k \Tilde{i}_{2k} \Tilde{j}_{2k}}$. 
            The number above the image is a sorted index correspond to Fig. \ref{fig:vfilter_downsampling_block_a}.]{
            \includegraphics[width=\textwidth]{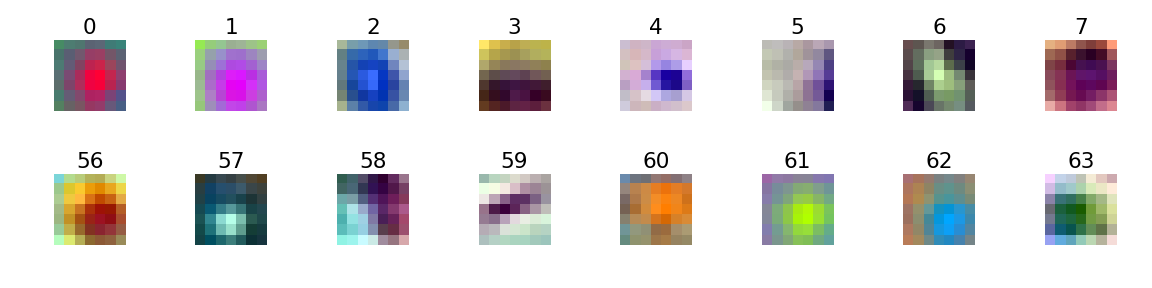}}
    \caption{Visualization a filter of down-sapmling shown at the bottom conv. layer of Fig. \ref{fig:downsampling_block} in ResNet34.}
    \label{fig:vfilter_downsampling_block}
\end{figure}

Teramoto \& Shouno propose a visualization method for the preferred stimulus as a convolution filter in the second layer of VGG \cite{teramoto2019vgg}.
Let $W^{l}_{pqij}$ be the convolutional weight connected from channel $q$ in layer $l$ to channel $p$ in layer $l + 1$ , and let $i$ and $j$ be spatial index.
Then, the method is to use the weight $\widetilde{W}^{2}_{p}$ as the $p$-th filter in the second layer.
The weight $\widetilde{W}^{2}_{p}$ is represented as 
\begin{equation}
    \widetilde{W}^{2}_{pqij} = \sum_{k} W^{2}_{pk \Tilde{i}_{pk} \Tilde{j}_{pk}} W^{1}_{kqij}
\end{equation}
where $(\Tilde{i}_{pk}, \Tilde{j}_{pk}) = \arg \max_{i', j'} | W^{2}_{pki'j'}|$.
In general, this method is an approximated visualization for filters in higher layers because CNNs have non-linear function between convolutional layers.
We call the filter to ``virtual filter'', and apply this method to the second down-sampling layer in ResNet34. 
Fig. \ref{fig:vfilter_downsampling_block} shows the virtual filters and the first filters in ResNet34.
Looking at the coupling coefficients of the filters in Fig. \ref{fig:vfilter_downsampling_block}, it can be seen that the coupling to similar filters is stronger.
ResNet with a skip structure also acquires features similar to the column structure which is a biological finding.

\subsection{Analysis of Preferred Stimulus in Receptive Fields}
We focus on the preferred stimuli, which activate a neuron in the ResNets with strongly positive, in the input data set. 
In order to find the preferred stimuli, We feed validation images $X$ of ImageNet to DCNNs at first.
After that, in each layer, we align the stimulus with descending order of activation value.
Let $\bm{r}_{i}$ be the receptive field of neuron $i$
and let $f_{i} \left( \bm{x} \left[ \bm{r}_{i} \right] \right)$ be the activation value of neuron $i$ on a given receptive field image.
Now, we can describe the mean preferred stimulus image on positive validation images $X^{+}$ as
\begin{equation}
    \overline{\bm{x}}^{i} = \frac{1}{N} \sum_{\bm{x} \in X^{+}} \bm{x} \left[ \bm{r}_{i} \right].
\end{equation}
The positive validation images are validation images which the neuron activate positive and
the images are represented by 
\begin{equation}
    X^{+} = \left\{ \bm{x} \in X \ | \ f_{i} \left( \bm{x} \left[ \bm{r}_{i} \right] \right) > 0 \right\}.
\end{equation}

\begin{figure}[tbp]
    \subfloat[Channel 18 in first max-pooling layer.
              The receptive field size is $11 \times 11$.\label{fig:all_res_sample_maxpool}]{
              \includegraphics[width=0.32\textwidth]{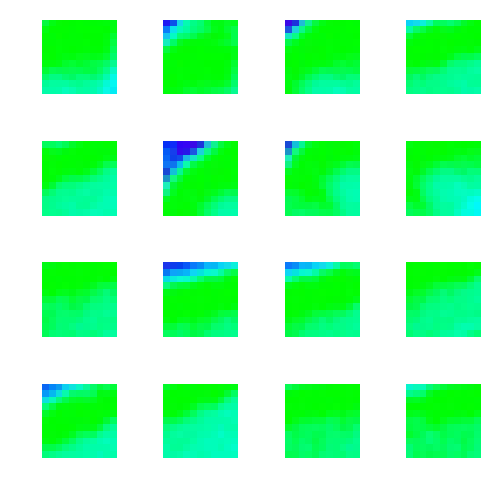}}
    \hfill
    \subfloat[Channel 18 in conv. layer in layer 3.
              The receptive field size is $27 \times 27$.\label{fig:all_res_sample_layer3}]{
              \includegraphics[width=0.32\textwidth]{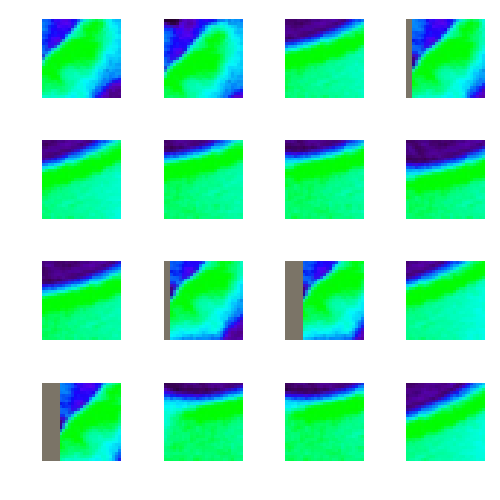}}
    \hfill
    \subfloat[Channel 18 in conv.layer in layer 7.
              The receptive field size is $59 \times 59$.\label{fig:all_res_sample_layer7}]{
              \includegraphics[width=0.32\textwidth]{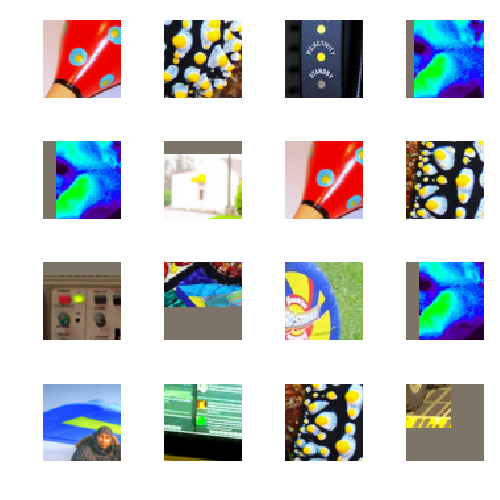}}
    \caption{Samples of the top 16 preferred stimulus images in ResNet34.
    }
    \label{fig:all_res_sample}
\end{figure}

\begin{figure}[tbp]
    \subfloat[First conv. filter of channel 18.\label{fig:all_res_f}]{
    \includegraphics[width=0.24\textwidth]{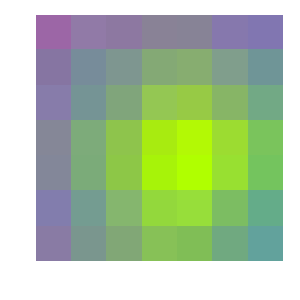}}
    \hfill
    \subfloat[Mean preferred stimulus image of channel 18 in first max-pooling layer.\label{fig:all_res_mrf_maxpool}]{
    \includegraphics[width=0.24\textwidth]{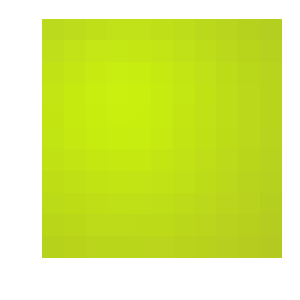}}
    \hfill
    \subfloat[Mean preferred stimulus image of channel 18 in conv.layer in layer 3.\label{fig:all_res_mrf_layer3}]{
    \includegraphics[width=0.24\textwidth]{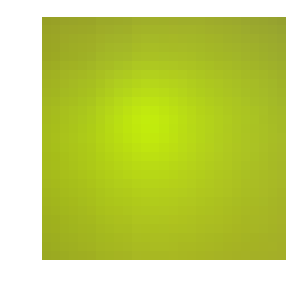}}
    \hfill
    \subfloat[Mean preferred stimulus image of channel 18 in conv. layer in layer 7.\label{fig:all_res_mrf_layer7}]{
    \includegraphics[width=0.24\textwidth]{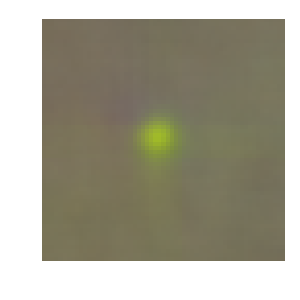}}
    \caption{First convolutional filter and mean preferred stimulus images in ResNet34.
    }
    \label{fig:all_res}
\end{figure}

\begin{figure}[tbp]
    \subfloat[Channel 19 in first max-pooling layer.
              \label{fig:all_plain_sample_maxpool}]{
              \includegraphics[width=0.32\textwidth]{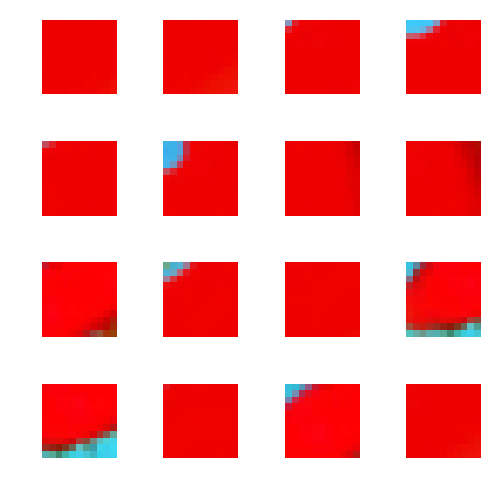}}
    \hfill
    \subfloat[Channel 19 in conv. layer in layer 3.
              \label{fig:all_plain_sample_layer3}]{
              \includegraphics[width=0.32\textwidth]{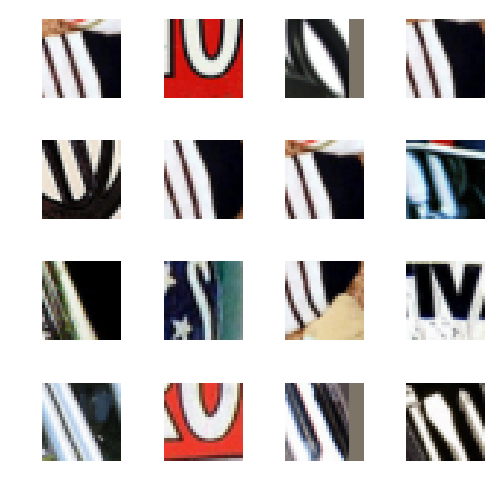}}
    \hfill
    \subfloat[Channel 19 in conv.layer in layer 7.
              \label{fig:all_plain_sample_layer7}]{
              \includegraphics[width=0.32\textwidth]{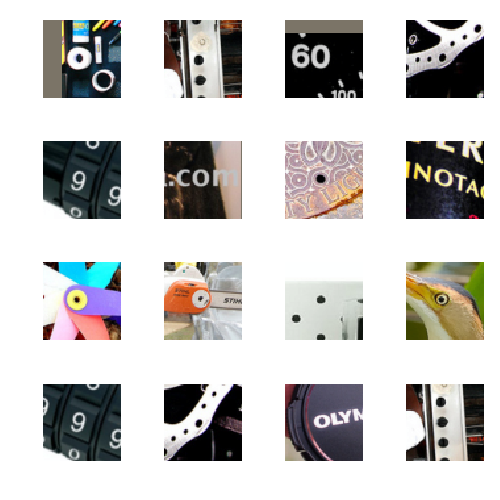}}
    \caption{Samples of the top 16 preferred stimulus images in PlainNet34.
    }
    \label{fig:all_plain_sample}
\end{figure}

\begin{figure}[tbp]
    \subfloat[First conv. filter of channel 19.\label{fig:all_plain_f}]{
    \includegraphics[width=0.24\textwidth]{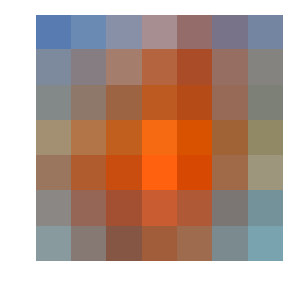}}
    \hfill
    \subfloat[Mean preferred stimulus image of channel 19 in first max-pooling layer.\label{fig:all_plain_mrf_maxpool}]{
    \includegraphics[width=0.24\textwidth]{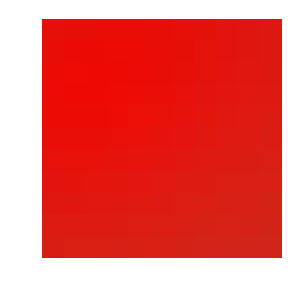}}
    \hfill
    \subfloat[Mean preferred stimulus image of channel 19 in conv.layer in layer 3.\label{fig:all_plain_mrf_layer3}]{
    \includegraphics[width=0.24\textwidth]{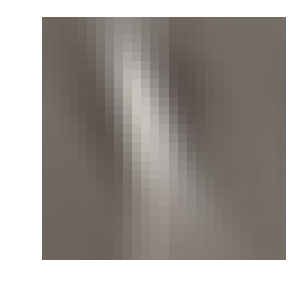}}
    \hfill
    \subfloat[Mean preferred stimulus image of channel 19 in conv. layer in layer 7.\label{fig:all_plain_mrf_layer7}]{
    \includegraphics[width=0.24\textwidth]{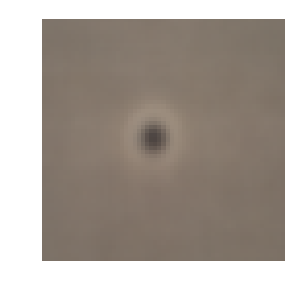}}
    \caption{First convolutional filter and mean preferred stimulus images in PlainNet34.
    }
    \label{fig:all_plain}
\end{figure}

We show a few examples of the top 16 at some neurons in Fig. \ref{fig:all_res_sample} and \ref{fig:all_plain_sample}, and the convolutional filter and the mean preferred stimulus images correspond to the neurons in Fig. \ref{fig:all_res} and \ref{fig:all_plain}.
We find that DCNNs prefer a variety features as higher layers from the sample of the preferred stimuli.
At first glance, Fig. \ref{fig:all_res_sample_layer7} and \ref{fig:all_plain_sample_layer7} appear to be an inconsistent sample, but there are central features from Fig. \ref{fig:all_res_mrf_layer7} and \ref{fig:all_plain_mrf_layer7}.
We find that 
the characteristics of the same channel are similar in different layers due to the skip connection of the ResNet.
We can see that
the mean preferred stimulus images can only find the broad tendencies but it is difficult to find the detailed properties of the neuron.

\subsection{Visualization Using Maximization Method}
\begin{figure}[tbp]
    \subfloat[One optimal input of channel 18 in first max-pooling layer.]{
    \includegraphics[width=0.32\textwidth]{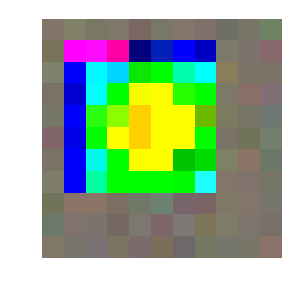}}
    \hfill
    \subfloat[One optimal input of channel 18 in conv. layer in layer 3.]{
    \includegraphics[width=0.32\textwidth]{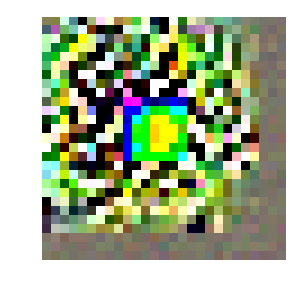}}
    \hfill
    \subfloat[One optimal input of channel 18 in conv. layer in layer 7.]{
    \includegraphics[width=0.32\textwidth]{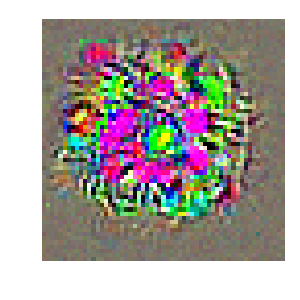}}
    \caption{Examples of visualizations by activation maximization for the neuron in ResNet34.}
    \label{fig:opt_neu_res}
\end{figure}

\begin{figure}[tbp]
    \subfloat[One optimal input of channel 18 in first max-pooling layer.\label{fig:opt_lay_res_maxpool}]{
    \includegraphics[width=0.32\textwidth]{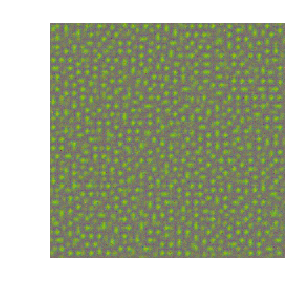}}
    \hfill
    \subfloat[One optimal input of channel 18 in conv. layer in layer 3.\label{fig:opt_lay_res_layer3}]{
    \includegraphics[width=0.32\textwidth]{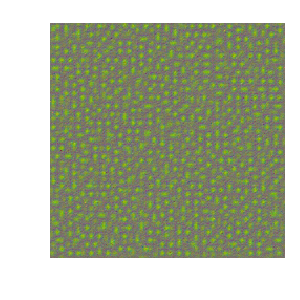}}
    \hfill
    \subfloat[One optimal input of channel 18 in conv. layer in layer 7.]{
    \includegraphics[width=0.32\textwidth]{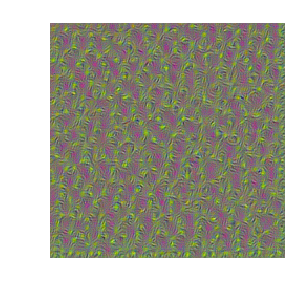}}
    \caption{Examples of visualizations by activation maximization for the channel in ResNet34. The image size is $224 \times 224$.}
    \label{fig:opt_lay_res}
\end{figure}

We apply activation maximization method\cite{erhan2009visualizing, simonyan2013deep} to ResNet34 and show the results for the neuron and the channel in the layer in Fig. \ref{fig:opt_neu_res} and \ref{fig:opt_lay_res}.
Optimizing for the neuron is to maximize the activation of the center neuron in a feature map.
and optimizing for the channel is to maximize the average of the activation of a channel.
We optimize the input by Adam optimizer \cite{kingma2014adam} with a learning rate of $0.1$ and a weight decay of $10^{-6}$.
In addition, we initialize the inputs from a zero image and iterate until $31$ times.

From the comparison of Fig. \ref{fig:all_res} and \ref{fig:opt_neu_res}, we can see that the results for optimizing for the neuron are similar to the results of the mean preferred stimulus images.
The visualization at higher layers reveals detailed properties for activation maximization, but only simple trends for mean  preferred stimulus images.
Especially, visualizing by activation maximization for the channel is a good-looking visualization of the neuron but the results vary according to various experimental conditions.

\subsection{Inactive Neurons}
For validation dataset images, we find that some channels in the first max-pooling layer have no output activation values in other words output zeros value because of ReLU activation function.
We call the channel to ``inactive neuron''.
In addition, we find that ResNet34 appears more inactive neurons rather than that of the PlainNet34 from Table \ref{tab:activation_count}.

To investigate the effect of the inactivate neuron on the classification,
we perform two classification experiments that add noise to the inactive neurons.
The one is to add noise to all inactive neurons
and the second is to add noise to one inactive neuron selected randomly every mini-batch.
We apply noise $\epsilon = \max(x, 0)$ where $x \sim \mathcal{N}(0, 1)$ to each spatial dimension of the inactive neuron.
Table \ref{tab:activation_count} shows the results,
$\Delta L$ means the value from all noised validation loss minus validation loss,
and $\Delta L_{rnd}$ means the value from randomly noised validation loss minus validation loss.
We can see that
the inactive neuron of ResNet34 effects classification task
because both $\Delta L$ and $\Delta L_{rnd}$ of ResNet34 are positive and bigger than that of PlainNet34.
\begin{table}[tbp]
    \centering
    \caption{
        Count of the inactive neurons and effect of the inactive neuron
        in first max-pooling layer for validation dataset
        in ResNet34 and PlainNet34.
    }
    \begin{tabular}{c|r|rr}
    \hline
Model & \# of inactive neurons & $\Delta L$  & $\Delta L_{rnd}$ \\ \hline
ResNet34   & $13$ &   $1.26962{\rm e}+0$ &  $ 2.40560{\rm e}-2$  \\
PlainNet34 &  $2$ &  $-1.66893{\rm e}-6$ &  $-8.34465{\rm e}-7$ \\ \hline
    \end{tabular}
    \label{tab:activation_count}
\end{table}

\section{Conclusion}
We perform analysis by using preferred stimulus and activation maximization to ResNets.
Using both methods, we can find that ResNet has orientation selective neurons and double opponent color neurons.
Both methods are able to characterize the lower layers well but it is harder to use the analysis for the higher layers.
We find that there are inactive neurons for the classification task in ResNet34.
We speculate that this phenomenon is due to channel sharing by skip connections.
One hypothesis is that some channels are used for features that are not similar to the features of first convolutional layers.
In future works, we need to consider methods that can perform analysis to the higher layers, and examine the evidence to support our hypothesis.

\bibliographystyle{splncs04}
\bibliography{ref}

\begin{thebibliography}{10}
\providecommand{\url}[1]{\texttt{#1}}
\providecommand{\urlprefix}{URL }
\providecommand{\doi}[1]{https://doi.org/#1}

\bibitem{imagenet_cvpr09}
Deng, J., Dong, W., Socher, R., Li, L.J., Li, K., Fei-Fei, L.: {ImageNet: A
  Large-Scale Hierarchical Image Database}. In: CVPR09 (2009)

\bibitem{erhan2009visualizing}
Erhan, D., Bengio, Y., Courville, A., Vincent, P.: Visualizing higher-layer
  features of a deep network. University of Montreal  \textbf{1341}(3), ~1
  (2009)

\bibitem{Fukushima1980}
Fukushima, K.: Neocognitron: A self-organizing neural network model for a
  mechanism of pattern recognition unaffected by shift in position. Biological
  Cybernetics  \textbf{36}(4),  193--202 (Apr 1980). \doi{10.1007/BF00344251},
  \url{https://doi.org/10.1007/BF00344251}

\bibitem{He2016ResNet}
He, K., Zhang, X., Ren, S., Sun, J.: Deep residual learning for image
  recognition. In: 2016 IEEE Conference on Computer Vision and Pattern
  Recognition (CVPR). pp. 770--778 (June 2016). \doi{10.1109/CVPR.2016.90}

\bibitem{hubel1959receptive}
Hubel, D.H., Wiesel, T.N.: Receptive fields of single neurones in the cat's
  striate cortex. The Journal of physiology  \textbf{148}(3),  574--591 (1959)

\bibitem{kingma2014adam}
Kingma, D.P., Ba, J.: Adam: A method for stochastic optimization. arXiv
  preprint arXiv:1412.6980  (2014)

\bibitem{AlexNet2012}
Krizhevsky, A., Sutskever, I., Hinton, G.E.: Imagenet classification with deep
  convolutional neural networks. In: Pereira, F., Burges, C.J.C., Bottou, L.,
  Weinberger, K.Q. (eds.) Advances in Neural Information Processing Systems 25,
  pp. 1097--1105. Curran Associates, Inc. (2012),
  \url{http://papers.nips.cc/paper/4824-imagenet-classification-with-deep-convolutional-neural-networks.pdf}

\bibitem{Lecun1998}
Lecun, Y., Bottou, L., Bengio, Y., Haffner, P.: Gradient-based learning applied
  to document recognition. In: Proceedings of the IEEE. pp. 2278--2324 (1998)

\bibitem{liao2016bridging}
Liao, Q., Poggio, T.: Bridging the gaps between residual learning, recurrent
  neural networks and visual cortex. arXiv preprint arXiv:1604.03640  (2016)

\bibitem{ILSVRC15}
Russakovsky, O., Deng, J., Su, H., Krause, J., Satheesh, S., Ma, S., Huang, Z.,
  Karpathy, A., Khosla, A., Bernstein, M., Berg, A.C., Fei-Fei, L.: {ImageNet
  Large Scale Visual Recognition Challenge}. International Journal of Computer
  Vision (IJCV)  \textbf{115}(3),  211--252 (2015).
  \doi{10.1007/s11263-015-0816-y}

\bibitem{selvaraju2017grad}
Selvaraju, R.R., Cogswell, M., Das, A., Vedantam, R., Parikh, D., Batra, D.:
  Grad-cam: Visual explanations from deep networks via gradient-based
  localization. In: 2017 IEEE International Conference on Computer Vision
  (ICCV). pp. 618--626. IEEE (2017)

\bibitem{simonyan2013deep}
Simonyan, K., Vedaldi, A., Zisserman, A.: Deep inside convolutional networks:
  Visualising image classification models and saliency maps (2013)

\bibitem{simonyan2014vgg}
Simonyan, K., Zisserman, A.: Very deep convolutional networks for large-scale
  image recognition. arXiv preprint arXiv:1409.1556  (2014)

\bibitem{springenberg2014striving}
Springenberg, J.T., Dosovitskiy, A., Brox, T., Riedmiller, M.: Striving for
  simplicity: The all convolutional net. arXiv preprint arXiv:1412.6806  (2014)

\bibitem{szegedy2015inception}
Szegedy, C., Liu, W., Jia, Y., Sermanet, P., Reed, S., Anguelov, D., Erhan, D.,
  Vanhoucke, V., Rabinovich, A.: Going deeper with convolutions. In:
  Proceedings of the IEEE conference on computer vision and pattern
  recognition. pp.~1--9 (2015)

\bibitem{teramoto2019vgg}
Teramoto, T., Shouno, H.: A study of inner feature continuity of the vgg model.
  In: IEICE Technical Report. pp. 239--244. IEICE (March 2019)

\bibitem{Yamins8619}
Yamins, D.L.K., Hong, H., Cadieu, C.F., Solomon, E.A., Seibert, D., DiCarlo,
  J.J.: Performance-optimized hierarchical models predict neural responses in
  higher visual cortex. Proceedings of the National Academy of Sciences
  \textbf{111}(23),  8619--8624 (2014). \doi{10.1073/pnas.1403112111},
  \url{https://www.pnas.org/content/111/23/8619}

\bibitem{zeiler2014visualizing}
Zeiler, M.D., Fergus, R.: Visualizing and understanding convolutional networks.
  In: European conference on computer vision. pp. 818--833. Springer (2014)

\end{thebibliography}
\end{document}